\documentclass[compsoc, conference, a4paper, 10pt, times]{IEEEtran}

\usepackage{cite}
\usepackage{amsmath,amssymb,amsfonts}
\usepackage{algorithmic}
\usepackage{graphicx}
\usepackage{textcomp}
\usepackage{xcolor}
\usepackage{booktabs}
\usepackage[hidelinks]{hyperref}
\usepackage{listings}

\definecolor{delim}{RGB}{20,105,176}
\definecolor{numb}{RGB}{106, 109, 32}
\definecolor{string}{rgb}{0,0,0}
\definecolor{celadon}{rgb}{0.67, 0.88, 0.69}

\lstdefinelanguage{json}{
    frame=single,
    numbers=left,
    numberstyle=\tiny\color{gray},
    stepnumber=1,
    rulecolor=\color{black},
    tabsize=2,
    showspaces=false,
    showtabs=false,
    showstringspaces=false,
    breaklines=true,
    postbreak=\raisebox{0ex}[0ex][0ex]{\ensuremath{\color{gray}\hookrightarrow\space}},
    breakatwhitespace=true,
    basicstyle=\ttfamily\tiny,
    upquote=true,
    morestring=[b]",
    stringstyle=\color{string},
    literate=
     *{0}{{{\color{numb}0}}}{1}
      {1}{{{\color{numb}1}}}{1}
      {2}{{{\color{numb}2}}}{1}
      {3}{{{\color{numb}3}}}{1}
      {4}{{{\color{numb}4}}}{1}
      {5}{{{\color{numb}5}}}{1}
      {6}{{{\color{numb}6}}}{1}
      {7}{{{\color{numb}7}}}{1}
      {8}{{{\color{numb}8}}}{1}
      {9}{{{\color{numb}9}}}{1}
      {\{}{{{\color{delim}{\{}}}}{1}
      {\}}{{{\color{delim}{\}}}}}{1}
      {[}{{{\color{delim}{[}}}}{1}
      {]}{{{\color{delim}{]}}}}{1},
}

\lstdefinelanguage{Markdown}{
    basicstyle=\ttfamily\tiny,
    breaklines=true,
    columns=fullflexible,
    upquote=true,
    sensitive=true,
    morecomment=[l]{\#},
    morecomment=[l]{>},
    morestring=[b]",
    morestring=[b]',
    moredelim=[s][\bfseries]{**}{**},
    moredelim=[s][\bfseries\itshape]{_}{_},
    moredelim=[s][\color{blue}]{`}{`},
}

\begin{document}

\title{Addressing Labelled Data Scarcity:\\Taxonomy-Agnostic Annotation of PII Values in HTTP Traffic using LLMs}

\author{\em Anonymous Authors}

\iftrue
\author{\IEEEauthorblockN{1\textsuperscript{st} Thomas Cory}
\IEEEauthorblockA{\textit{Technische Universität Berlin} \\
Berlin, Germany \\
cory@tu-berlin.de}
\and
\IEEEauthorblockN{2\textsuperscript{nd} Axel Küpper}
\IEEEauthorblockA{\textit{Technische Universität Berlin} \\
Berlin, Germany \\
axel.kuepper@tu-berlin.de}
}
\fi

\maketitle

\begin{abstract}
Automated privacy audits of web and mobile applications often analyse outbound HTTP traffic to detect Personally Identifiable Information (PII) leakage. However, existing learning-based detectors typically depend on scarce, manually labelled traffic and are tightly coupled to fixed label taxonomies, limiting transferability across domains and evolving definitions of PII. This paper investigates whether Large Language Models (LLMs) can support \textit{taxonomy-agnostic} annotation of explicitly transmitted PII values in HTTP message bodies when the taxonomy is provided at runtime. We introduce a multi-stage LLM-based pipeline that combines deterministic pre-processing with label-level classification, targeted instance-level value annotation, and output validation. To enable controlled evaluation and exemplar-based prompting without relying on sensitive real-user captures, we further propose an LLM-based generator for synthetic HTTP traffic with manually validated, taxonomy-derived PII annotations. We evaluate the approach across three taxonomies spanning different PII domains and granularity levels. Results show that the pipeline accurately detects PII types and extracts corresponding values for concrete PII taxonomies. Overall, our findings position LLMs as a promising foundation for flexible, taxonomy-agnostic traffic annotation and for creating labelled data under evolving privacy taxonomies.
\end{abstract}



\section{Introduction}
\label{sec:introduction}

Privacy regulations, such as the GDPR~\cite{gdpr2016} and CCPA~\cite{ccpa2018}, introduce comprehensive legal frameworks regulating the collection and processing of Personally Identifiable Information (PII). However, assessing whether the data and privacy practices of web and mobile applications are compliant with these regulatory frameworks is challenging and often involves time-consuming manual input. The last decade has seen repeated attempts to reduce the manual overhead of such privacy audits through the development of automated approaches. Automated analysis of applications' network traffic plays a key role in these efforts, as network traffic represents the primary vector for PII leakage, i.e. the transfer of PII out of the control sphere of the data subject~\cite{ReCon2016}. Early attempts employed keyword-based heuristics to detect PII values in transmissions, but were inherently limited due to the variability of PII values. Later approaches employed machine-learning techniques such as Support Vector Machines (SVMs) or Bidirectional Encoder Representations from Transformers (BERT)~\cite{grancharova2021applying} to detect PII values with greater flexibility. 

However, these approaches still have a key drawback: they require training on high-quality labelled data, which traditionally requires careful manual effort to create. Furthermore, they are locked to the label taxonomy and type of data they were trained on, greatly limiting their transferability, as even small changes to either generally negatively impact their performance substantially. This is particularly problematic for the task of PII detection, as the definition of an exhaustive taxonomy of PII types is challenging. Instead, existing taxonomies focus on specific domains or granularity levels and are rarely fully compatible with each other~\cite{szawerna2024pseudonymization}. This challenge is reflected in the PII detection and de-identification literature more broadly: recent work reports little uniformity in how researchers and corpus creators classify PII, with taxonomies differing markedly in granularity and semantic scope, which in turn affects both detection and labelling behaviour \cite{szawerna2025devil}. Furthermore, the task of manually labelling and curating ground-truth datasets is highly labour-intensive and needs to be repeated for every update made to a label taxonomy~\cite{szawerna2024pseudonymization, CoryPoPETs2026}. The result is a widespread scarcity of high-quality datasets of labelled network traffic with up-to-date label taxonomies and limited flexibility of existing PII detectors.

Generative models, particularly Large Language Models (LLMs), represent a paradigm shift in this regard: rather than requiring careful training and attunement to specific label taxonomies and input types, they are generalists that can be prompted to perform a variety of tasks without requiring retraining~\cite{yang2023exploring}. Prior work has demonstrated their applicability for a wide range of language processing and specifically Named Entity Recognition (NER) and annotation tasks~\cite{GPTNER2025, CoryPoPETs2026}, motivating us to evaluate their suitability for taxonomy-agnostic annotation of PII values in HTTP traffic. 

In contrast to prior network-traffic PII detectors that are either heuristic-driven or trained and evaluated on fixed label sets, our work evaluates whether LLMs can mitigate labelled-data scarcity in network-traffic privacy auditing by enabling taxonomy-agnostic, value-level annotation of PII in HTTP messages and by supporting controlled generation of labelled synthetic traffic. We address the following research questions:

\begin{enumerate}
    \item[\textbf{RQ1:}] To what extent can an LLM-based pipeline accurately identify and annotate explicitly transmitted PII values in HTTP message payloads when the label taxonomy is provided at runtime?
    \item[\textbf{RQ2:}] How do key architectural choices affect label-level and instance-level annotation quality?
    \item[\textbf{RQ3:}] To what extent can synthetic HTTP messages with ground-truth PII annotations support evaluation and exemplar-based prompting, and how robust is performance across taxonomies that differ in domain focus and granularity?
\end{enumerate}

In addressing these research questions, our contributions are threefold:

\begin{enumerate}
    \item[\textbf{C1:}] We present a multi-stage LLM-based approach for the highly granular identification and annotation of PII entities in HTTP message payloads, capable of switching label taxonomies at runtime.
    \item[\textbf{C2:}] We complement the presented annotation approach with an LLM-based approach for the generation of synthetic HTTP traffic with pre-labelled PII values, reducing the need for labour-intensive manual annotation and curation of ground-truth datasets for shifting label taxonomies.
    \item[\textbf{C3:}] We evaluate the presented annotation approach on three PII label taxonomies covering different domains and granularity levels to assess the performance, suitability, and flexibility of LLM-based PII value annotation.
\end{enumerate}

\section{Related Work}
\label{sec:related_work}

A substantial body of prior work has studied privacy leakage by analysing outbound network traffic from mobile applications. Early systems relied on rule- and keyword-based heuristics to identify sensitive values such as device identifiers or locations within HTTP requests, often via VPN-based interception on unrooted devices~\cite{PrivacyGuard2015,AntMonitor2015,Haystack2016}. While effective for well-structured payloads and known formats, such heuristics are brittle under variable value representations, application-layer encoding, and evolving schemas. Later systems therefore introduced learning-based PII detectors. ReCon~\cite{ReCon2016} interposes on network traffic and uses machine learning to detect and extract PII values across platforms and apps, while PrivacyProxy~\cite{PrivacyProxy2017} identifies likely PII through app-specific signatures and crowd aggregation, reducing raw-traffic exposure through hashed representations. Ren et al.~\cite{RenNDSS2018} further use ReCon longitudinally, showing that PII leakage patterns change across app versions and that transport-layer protection is adopted slowly. More recently, Kohli et al.~\cite{kohli2025learning} compare tabular models over structured packet features with document-classification models over raw packet content, including transformer-based architectures for mobile network packets. This line of work establishes network traffic as a primary vantage point for privacy auditing, but remains largely tied to the label schemes, feature representations, and training data available at development time.

LLMs have also been explored for structured and unstructured information extraction, including NER and privacy-related labelling. GPT-NER~\cite{GPTNER2025} reformulates NER as a generation task with explicit boundary markers and self-verification, achieving strong performance particularly in low-resource regimes. In privacy settings, Cory et al. (2026)~\cite{CoryPoPETs2026} present a modular LLM-based pipeline for fine-grained, word-level annotation of GDPR transparency requirements in privacy policies, combining passage-level classification, retrieval-augmented generation, and self-correction. Zeng et al.~\cite{PrivacyAnnotation2025} similarly propose a staged LLM-driven pipeline for annotating privacy phrases in user--LLM interactions, showing that decomposition can outperform direct prompting. Shen et al.~\cite{shen2025pii} further highlight the need for fine-grained and context-aware privacy reasoning, as systems that detect PII may still struggle to determine whether it is relevant to a given query. Meena et al.~\cite{MultilingualPII2025} complement these approaches with a human-in-the-loop framework for multilingual PII annotation, emphasising guidelines and quality assurance for fine-grained and locale-specific labels. Together, these works motivate LLM-based privacy annotation, while highlighting the importance of task decomposition, schema adherence, contextual disambiguation, and quality control.

Synthetic data generation has long been used to support modelling under privacy, access, and resource constraints. CTGAN~\cite{CTGAN2019} demonstrates high-fidelity generation for mixed-type tabular data, medGAN~\cite{medGAN2017} synthesises high-dimensional discrete patient records with promising downstream utility, and PATE-GAN~\cite{PATEGAN2019} frames synthetic data generation as a mechanism for modelling with differential-privacy guarantees. However, studies comparing synthetic and real data caution against treating synthetic data as a direct substitute for real-world observations. Reiner Benaim et al.~\cite{reiner2020analyzing} compare synthetic and real medical datasets across five observational studies and show that utility depends on study design, sample size, and data complexity, while Schmidhuber and Kruschwitz ~\cite{kruschwitz-schmidhuber-2024-llm} find that LLM-generated synthetic data can support toxicity detection in some settings but not uniformly across tasks and datasets. Synthetic data has also been applied directly to privacy-sensitive information: Savkin et al.~\cite{savkin2025spy} introduce SPY, an LLM-generated synthetic benchmark for PII detection that emulates realistic PII-containing scenarios without exposing real personal data, and Löbner et al.~\cite{lobner2025mitigating} use synthetic CV data for de-identifying privacy-sensitive information in recruitment. These works motivate our use of synthetic HTTP traffic with ground-truth PII annotations to alleviate labelled-data scarcity and enable controlled, taxonomy-specific evaluation, while also underscoring the need to validate synthetic results against real-world traffic.
\begin{figure*}[t]
  \centering
  \includegraphics[width=0.9\linewidth]{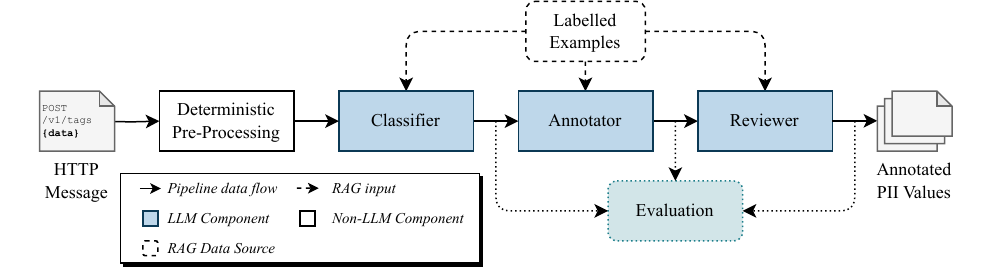}
  \caption{Overview of the proposed multi-stage annotation pipeline. The pipeline normalises HTTP message bodies, performs label-level classification followed by targeted value annotation, and applies a review stage to refine and complete the resulting annotations.}
  \label{fig:annotation_pipeline}
\end{figure*}

\section{Annotation Pipeline}
\label{sec:annotation_pipeline}

We implement PII annotation in HTTP traffic as a multi-stage pipeline that decomposes the overall task into specialised sub-problems. This design follows the observation that jointly inferring \textit{what} types of PII occur in a message and \textit{where} their corresponding values begin and end is error-prone in network traffic, where short identifiers, numeric values, and encoding artefacts are common. We base our pipeline architecture on the multi-stage annotation approach presented by Cory et al. (2026)~\cite{CoryPoPETs2026}. Figure~\ref{fig:annotation_pipeline} summarises our pipeline, which consists of (i) deterministic pre-processing to normalise message bodies, (ii) a two-stage annotation procedure comprising label-level classification and instance-level value annotation, and (iii) a review layer that performs targeted correction and completion.

\subsection{Pre-processing}
\label{subsec:pre_processing}

Prior to PII detection, HTTP message bodies are normalised to mitigate encoding-related variability and to present downstream stages with a consistent, human-readable representation. We first apply URL decoding (RFC~3986), using \texttt{unquote\_plus} to resolve percent-encoded sequences (e.g., \texttt{\%2B}) and to map plus signs to spaces where applicable. We then resolve HTML character references, including named entities (e.g., \texttt{\&amp;}) and numeric references (e.g., \texttt{\&\#x40;}), by decoding them to their corresponding Unicode characters.

In addition to these canonical transformations, HTTP payloads may embed values that are encoded at the application layer (e.g., Base64). To accommodate this, we apply an optional decoding pass for common encodings when there is strong evidence that a substring represents an encoded payload. Concretely, we attempt Base64 decoding only for substrings that match a conservative syntactic pattern and decode to predominantly printable UTF-8 text; otherwise, the original representation is retained. To avoid information loss and to ensure traceability, we keep both the original body and its normalised (and, where applicable, decoded) representation, and we map annotated spans back to the original message when reporting annotations.

\subsection{Two-stage annotation}
\label{subsec:two_stage_annotation}

We perform annotation in two stages: label-level classification followed by instance-level value extraction. Let $m$ denote an HTTP message body and $\mathcal{L}$ the label taxonomy. The classifier predicts a set of labels $L_m \subseteq \mathcal{L}$ that are explicitly present in $m$. The extractor then focuses on identifying the concrete value instances for the predicted labels, yielding the final annotation set. We summarise this decomposition as:
\begin{equation}
    \text{Annotate}(m) = \text{Extract}\!\left(m,\, \text{Classify}(m)\right).
\end{equation}

This separation is motivated by the structure of HTTP traffic. Many messages contain heterogeneous key-value pairs and nested serialisations (e.g., JSON), where the same surface form can correspond to multiple PII types depending on context. By conditioning extraction on the (typically small) set of predicted labels, the extraction stage can allocate its capacity to boundary detection and disambiguation rather than simultaneously exploring the full taxonomy.

The classification stage operates under a strict \textit{explicit-value} requirement: a label is emitted only if at least one corresponding value is present in the message body. In particular, we avoid inferring PII from key names alone (e.g., a field name such as \texttt{email} with an empty value), and we do not attempt to guess or reconstruct values that are not explicitly transmitted.

Given $m$ and $L_m$, the extraction stage performs targeted named-entity recognition over the normalised body, producing a set of annotation objects of the form \texttt{(data\_type, value)}, where \texttt{data\_type} is a label in $\mathcal{L}$ and \texttt{value} is the extracted string. Multiple instances per label are permitted when distinct values occur within the same message (e.g., multiple names in a registration form). Ambiguity is addressed through context-based disambiguation using cues available in network traffic. Key names provide strong priors (e.g., \texttt{age}, \texttt{weight}, \texttt{email}), while value patterns (e.g., email address structure, international phone prefixes) and local formatting (e.g., units such as \texttt{kg} or \texttt{cm}, locale-specific separators) further constrain interpretation. Where payloads contain nested structures, the structural position and neighbouring fields provide additional semantic context that helps distinguish short identifiers and numeric values that would otherwise be underspecified.

\subsection{Post-annotation Review}
\label{subsec:review}

Finally, we apply a review stage that implements a targeted quality assurance mechanism analogous to adjudication in manual annotation workflows. Rather than annotating a message independently, the reviewer is presented with the original message body and the initial annotation set and is tasked with refining the annotations. In practice, this stage primarily addresses three recurrent error modes in network-traffic annotation: missed values (false negatives), incorrect label assignments arising from ambiguity, and boundary errors where extracted spans are truncated or include surrounding delimiters. The reviewer may therefore add missing annotations, correct labels based on the message context and taxonomy definitions, and adjust value boundaries. By explicitly conditioning review on the prior output, this stage focuses effort on correction and completion, with the goal of yielding more consistent annotations without requiring a second full pass from scratch.
\section{LLM Inference Harness}
\label{sec:llm_harness}

\begin{figure}[th]
  \centering
  \includegraphics[width=\linewidth]{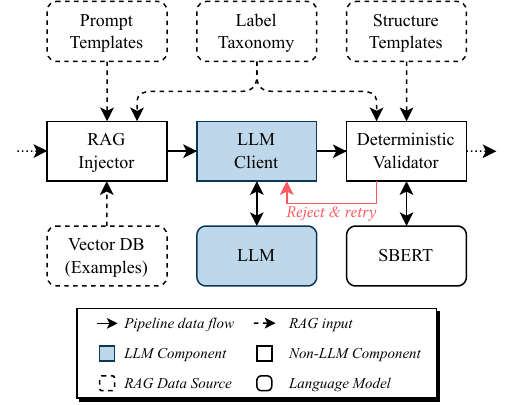}
  \caption{Overview of the LLM harness used within each pipeline component. The harness enriches inputs with retrieved, task-relevant exemplars and applies task-specific output validation to ensure schema and taxonomy compliance.}
  \label{fig:llm_harness}
\end{figure}

Each LLM-based component in our pipeline is wrapped by a shared harness that enriches the model input with task-relevant context via retrieval-augmented generation (RAG), and validates and normalises the model output before it is passed to subsequent stages. This harness, illustrated in Figure~\ref{fig:llm_harness}, addresses two recurrent failure modes in LLM-driven extraction pipelines: limited adherence to domain-specific label vocabularies and occasional violations of the expected output structure.

\subsection{Prompt preparation}
\label{subsec:prompt_engineering}

We follow a structured prompt engineering methodology tailored to information extraction under domain-specific taxonomies. First, we explicitly ground each prompt in the target taxonomy and require verbatim label copying to reduce label paraphrasing and out-of-vocabulary predictions. Where supported, we enforce a fixed JSON schema through structured output interfaces and reiterate exact string-matching requirements for label names. Second, to improve span-level extraction fidelity, we specify boundary rules that favour minimal spans (excluding syntactic delimiters such as quotes and separators) and treat composite fields as sets of atomic values where values are explicitly separated in the payload. Finally, we provide guidance for ambiguity resolution in network traffic, where short strings and numeric values are frequently underspecified without context. In these cases, the prompts instruct the model to rely on contextual cues such as key names, units, and structural position, and to omit an annotation when the available evidence does not support a confident assignment under the taxonomy definitions.

Across tasks, we employ a consistent prompt structure that combines role priming with explicit non-goals and lightweight self-checks. Role priming constrains the expected behaviour (e.g., classifier vs. annotator), while explicit non-goals prevent undesirable transformations (e.g., normalising values or inferring missing information). Short positive and negative examples are used selectively to illustrate boundary cases and to demonstrate the precise output format. Representative of all system prompts used in our approach, the annotator's system prompt is provided in Appendix~\ref{app:app:prompt_annotator}.

To improve robustness across heterogeneous HTTP payload structures and promote taxonomy compliance, we augment each model call with a small set of retrieved examples. We combine two complementary retrieval signals. The first targets \emph{structural similarity}, retrieving examples with comparable serialisation formats, key-value layouts, or schema fragments. The second targets \emph{label coverage}, ensuring that the retrieved context includes at least one demonstration for each label predicted to be present in the input (where applicable).

For similarity-based retrieval, we maintain a curated database of annotated examples $\mathcal{E} = \{(m_i, A_i)\}_{i=1}^{N}$, where $m_i$ denotes a message body and $A_i$ its ground-truth annotations. We index messages using sentence-level embeddings and retrieve the top-$k$ nearest neighbours under cosine similarity. In practice, this yields examples that share relevant structural features with the input (e.g., comparable JSON nesting or field naming conventions), which in turn improves output formatting and boundary selection. For label-aware retrieval, we select a subset of examples that collectively cover the target label set. To minimise the number of required examples, we implement this as a greedy selection procedure over $\mathcal{E}$, adding examples that maximise coverage of labels not yet observed in the retrieved context. This mechanism is particularly valuable when the input or taxonomy contains uncommon PII types.

Empirically, this combined RAG strategy serves three purposes. It anchors the model to the exact label vocabulary and output structure expected by the downstream validator, it enables lightweight domain adaptation by swapping or extending the database of examples without fine-tuning, and it mitigates cold-start effects for newly introduced labels by allowing support to be added through additional demonstrations.

\subsection{Output validation and normalisation}
\label{subsec:llm_output_validation}

\begin{figure*}[th]
  \centering
  \includegraphics[width=0.9\linewidth]{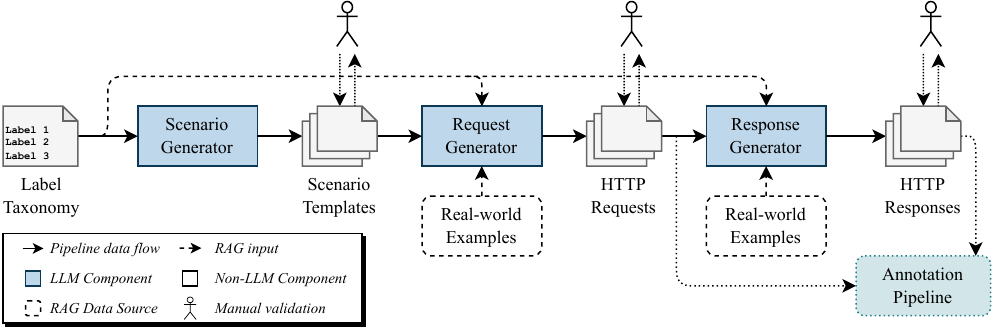}
  \caption{Overview of the synthetic HTTP message generator. Starting from a label taxonomy, the pipeline generates scenario templates, instantiates synthetic requests, and (optionally) synthesises corresponding responses. Each stage is followed by validation to enforce taxonomy adherence and to ensure that generated messages contain explicit, annotated PII values.}
  \label{fig:synthetic_request_generator}
\end{figure*}

To prevent error propagation, all LLM outputs are processed by a task-specific validation layer prior to downstream consumption. The validator first checks structural well-formedness (e.g., JSON parseability and schema conformance) and enforces taxonomy compliance by verifying that all emitted labels occur in the target taxonomy $\mathcal{L}$. When invalid labels occur, we apply a conservative normalisation step that maps common surface-form variations to valid taxonomy entries. Concretely, we embed the invalid label string and select the nearest neighbour among valid labels under cosine similarity:
\begin{equation}
    \text{correct}(l) = \arg\max_{l' \in \mathcal{L}} \cos(\phi(l), \phi(l')) ,
\end{equation}
where $\phi(\cdot)$ denotes the embedding function used for label strings. This mechanism resolves frequent benign deviations such as differences in capitalisation, hyphenation, minor paraphrases, or lightweight elaborations (e.g., ``E-mail'' versus ``Email'', or ``Date of Birth'' versus ``Date of birth''). To preserve precision, the correction is applied only when the similarity exceeds a predefined threshold; otherwise, the label is rejected and the corresponding annotation is discarded or flagged for re-querying, depending on the component. This design ensures that downstream stages receive outputs that are both structurally valid and consistent with $\mathcal{L}$.

\section{Synthetic HTTP Message Generator}
\label{sec:synthetic_http_message_generator}

In addition to our taxonomy-agnostic annotation pipeline, we introduce a complementary mechanism for producing labelled data at scale: a synthetic HTTP message generator that creates realistic request--response traffic containing \emph{explicit} PII values together with ground-truth annotations. Figure~\ref{fig:synthetic_request_generator} summarises the pipeline. The generator is designed to address a central bottleneck in supervised PII detection for network traffic, namely the limited availability of high-quality labelled datasets and the sensitivity of real traffic to privacy and access constraints. By producing synthetic traffic that follows plausible application scenarios while avoiding the use of real user data, the generator supports controlled training, ablation, and evaluation studies under labelled-data scarcity.

The generator is taxonomy-driven and LLM-agnostic: it consumes a label taxonomy (e.g., derived from platform disclosures or domain-specific PII schemes) and interfaces with any OpenAI-compatible API, enabling use with both hosted and locally deployed models. The pipeline proceeds in three stages: scenario generation, message generation, and validation. An optional third generation stage synthesises HTTP responses conditioned on previously generated requests, yielding bidirectional request--response traces suitable for end-to-end evaluation.

\subsection{Scenario and message generation}
\label{subsec:scenario_generation}

The first stage constructs \emph{scenario templates} that specify families of semantically related HTTP interactions. Given a taxonomy $\mathcal{L}$, the LLM produces a set of scenarios that jointly aim to cover the taxonomy while remaining consistent with common application flows (e.g., account creation, session management, checkout, telemetry submission). Each scenario template includes a goal description, host and endpoint variants, plausible HTTP methods and content types, and a distribution over candidate PII types with typical placements (path, query, headers, body). The candidate distribution is used to bias subsequent instantiation towards scenarios that naturally elicit particular labels, thereby improving coverage without requiring manual scenario design.

In the second stage, the LLM instantiates each validated scenario into a set of concrete HTTP requests. Each generated request comprises a structured representation of the HTTP message (method, path, headers, and body) together with a list of PII annotations. An annotation specifies the taxonomy label, the placement within the request (path, query, header, or body), and the verbatim value string. Importantly, generated values are required to be explicitly present in the message; the generator does not rely on implicit attributes or placeholders that are absent from the payload. This mirrors the explicit-value constraint adopted in the annotation pipeline and ensures that the synthetic dataset is suitable for training and evaluation of value-level PII detectors.

Where enabled, the generator subsequently produces a corresponding HTTP response for each request. Response synthesis is conditioned on the request content and scenario context, resulting in responses that remain plausible (e.g., status codes, response bodies, and headers consistent with the preceding request) and that may contain additional annotated PII values (e.g., account identifiers returned upon registration). Responses are represented in the same structured format as requests and are validated under the same explicit-value requirement.

\subsection{Validation and coverage tracking}
\label{subsec:validation_and_coverage_tracking}

The generator incorporates validation after each generation stage to prevent low-quality outputs from entering the final dataset. For requests and responses, the validator checks well-formedness of the HTTP structure, verifies that all labels in the annotation list occur in $\mathcal{L}$, and enforces that each annotated value appears verbatim in the specified message component. Outputs that violate these constraints are rejected and may be regenerated depending on the pipeline configuration.

In addition, the pipeline tracks label coverage throughout generation. Concretely, it monitors which labels from $\mathcal{L}$ appear across validated scenarios and across validated messages, enabling users to quantify progress towards comprehensive coverage and to identify underrepresented types. In practice, perfect coverage is not always attainable for arbitrary taxonomies, as some labels may be difficult to embed realistically into common HTTP exchanges without contrived scenarios. Coverage tracking therefore serves as a diagnostic and control signal: it allows the generation process to be steered (e.g., by increasing the number of scenarios or requests per scenario, or by enriching the scenario set) until the desired balance between realism, diversity, and taxonomy representation is achieved.
\section{Evaluation}
\label{sec:evaluation}

This section empirically assesses the effectiveness and flexibility of our LLM-based pipeline for taxonomy-agnostic PII value annotation in HTTP traffic. We focus on two questions: (i) how accurately the pipeline identifies which PII types occur in a message and annotates their corresponding values, and (ii) how performance changes across pipeline configurations and across taxonomies that differ in domain focus and granularity.

\subsection{Methodology}
\label{subsec:methodology}

We evaluate the proposed annotation pipeline using a staged framework that supports component-level ablation and reports performance at two complementary granularities. First, a \textit{label-level} evaluation measures whether a pipeline configuration correctly identifies which PII types occur in a message. Second, an \textit{instance-level} evaluation assesses whether the pipeline annotates the corresponding values with correct boundaries. Unless stated otherwise, we report precision, recall, and $F_{1}$, computed from true positives (TP), false positives (FP), and false negatives (FN) under the respective matching criterion.

\subsubsection{Label-level evaluation}
\label{subsubsec:label_level_evaluation}

For each HTTP message, we derive the set of unique labels present in the ground truth annotations, $G$, and the set predicted by the pipeline, $P$. We then treat label detection as a multi-label presence task: a label counts as a true positive if it appears in both sets, as a false positive if it appears only in $P$, and as a false negative if it appears only in $G$. This formulation intentionally abstracts away from within-message frequency and evaluates whether the pipeline detects the presence of each PII type at least once in the message body.

\subsubsection{Instance-level evaluation}
\label{subsubsec:instance_level_evaluation}

Instance-level evaluation considers annotated values. We represent each annotated instance as a tuple $(l, v)$ consisting of a label $l$ and its string value $v$. We report two complementary matching regimes.

Under \textit{exact} matching, a predicted instance is counted as correct only if both the label and the value string match the ground truth exactly. This strict criterion directly captures boundary errors and normalisation artefacts, and is therefore informative for downstream use cases that require verbatim leakage identification.

To account for minor, non-semantic deviations in annotation (e.g., whitespace differences, encoding-related character variants, or small boundary shifts), we additionally report a \textit{fuzzy} matching score. Concretely, for each ground-truth instance $(l, v_g)$, we consider predicted instances with the same label and select the candidate with the highest string similarity (computed using the \texttt{SequenceMatcher} algorithm). A match is accepted if the similarity exceeds a threshold $\tau$. Following prior work~\cite{CoryPoPETs2026}, we set $\tau = 0.8$. To prevent double-counting, we apply greedy one-to-one assignment: once a predicted instance has been matched to a ground-truth instance, it is removed from further consideration. FP and FN are derived from the remaining unmatched predictions and ground-truth instances, respectively.

\subsubsection{Ablation protocol}
\label{subsubsec:ablation_analysis}

We evaluate two configurations. In the \textit{two-stage} configuration, the classifier predicts the set of labels present in a message, and the annotator is constrained to extract values only for these predicted labels; the reviewer then refines the output. In the \textit{single-stage} configuration, the annotator receives only the raw message body and performs joint detection and annotation; the reviewer again refines the output. These configurations isolate the effect of the explicit decomposition introduced in Section~\ref{subsec:two_stage_annotation}.

To quantify the contribution of individual pipeline components, we evaluate outputs at multiple points in the pipeline. Specifically, we measure (i) label-level performance after the classification stage, (ii) instance-level performance after two-stage annotation, (iii) instance-level performance under a single-stage configuration in which the annotator receives only the raw message body and performs joint detection and annotation, and (iv) instance-level performance after the review stage. This design allows us to determine whether errors arise from missed type detection, from boundary selection during annotation, or from insufficient correction during review.

\subsection{Experimental setup}
\label{subsec:experiment_setup}

\begin{table*}[th]
\centering
\caption{Label- and instance-level $F_{1}$ scores across pipeline checkpoints for each taxonomy.}
\label{tab:evaluation_summary}
\small
\begin{tabular}{l|ccc|ccc|ccc}
\toprule
 &
\multicolumn{3}{|c}{\textit{AI4Privacy}} &
\multicolumn{3}{|c|}{\textit{mHealth}} &
\multicolumn{3}{c}{\textit{PlayStore}} \\
\midrule
Step & $F_{1 \text{label}}$ & $F_{1 \text{fuzzy}}$ & $F_{1 \text{exact}}$ &
$F_{1 \text{label}}$ & $F_{1 \text{fuzzy}}$ & $F_{1 \text{exact}}$ &
$F_{1 \text{label}}$ & $F_{1 \text{fuzzy}}$ & $F_{1 \text{exact}}$ \\
\midrule
Classifier               & 0.9860 & --     & --     & 0.9858 & --     & --     & 0.9109 & --     & -- \\
Annotator (single-stage) & 0.9857 & 0.9720 & 0.9377 & 0.9843 & 0.9593 & 0.9549 & 0.9091 & \textbf{0.8633} & \textbf{0.8061} \\
Annotator (two-stage)    & \textbf{0.9888} & \textbf{0.9776} & \textbf{0.9438} & 0.9880 & \textbf{0.9703} & \textbf{0.9658} & \textbf{0.9184} & 0.8625 & 0.8012 \\
Reviewer (single-stage)  & 0.9855 & 0.9674 & 0.9319 & 0.9870 & 0.9599 & 0.9548 & 0.9053 & 0.8570 & 0.8017 \\
Reviewer (two-stage)     & 0.9876 & 0.9747 & 0.9400 & \textbf{0.9881} & 0.9647 & 0.9597 & 0.9078 & 0.8554 & 0.7973 \\
\bottomrule
\end{tabular}
\end{table*}

We evaluate on three datasets, each associated with a distinct PII taxonomy that differs in domain focus and granularity. This choice is deliberate: our objective is to assess whether a single pipeline can be applied to heterogeneous label schemes provided at runtime, without retraining or taxonomy-specific engineering.

The first taxonomy, \textit{AI4Privacy}, is derived from the AI4Privacy \textit{pii-masking-200k}~\cite{ai4privacy200k} label set applied by Manietti and Elia~\cite{mainetti2025detecting}. Comprising 53 classes, it reflects a broad, natural-language-oriented PII scheme covering common identifiers (e.g., names, email addresses, phone numbers), financial attributes, government-issued identifiers, health-related attributes, biometrics, and other sensitive categories. The second taxonomy, \textit{mHealth}, corresponds to the Protected Health Information (PHI) scheme for mHealth apps presented by Cory et al. (2024)~\cite{CoryIWPE2024}. This taxonomy comprises 38 classes covering health-specific entities (e.g., vital signs and fitness metrics) alongside device identifiers and location-related attributes. The third taxonomy, \textit{PlayStore}, follows the Google Play Data Safety labels~\cite{datasafety}, which provide a higher-level categorisation of data types disclosed by Android applications across 38 classes (e.g., personal information, financial data, health and fitness, messages, media, app activity, browsing history, and device identifiers). The full label taxonomies are provided in Appendix~\ref{app:label_taxonomies}.

\subsubsection{Data preparation}
\label{subsubsec:data_preparation}

For each taxonomy, we use the synthetic HTTP message generator described in Section~\ref{sec:synthetic_http_message_generator} to create a labelled corpus of HTTP requests. Although the annotation pipeline can process both requests and responses, we restrict the evaluation to requests because the primary motivation underpinning our work is to assess PII leakage in outbound transmissions. To ensure reliable ground truth, we manually validate outputs throughout the generation pipeline, including scenario definitions, generated requests, and associated annotations. In particular, we check structural well-formedness, taxonomy compliance, verbatim occurrence of annotated values, and the semantic plausibility of label assignments.

To keep manual validation feasible while maintaining broad taxonomy coverage, we limit generation to 100 HTTP requests per scenario. The number of scenarios differs across taxonomies because the taxonomies vary in size, granularity, and semantic scope. The resulting corpora comprise 35 scenarios for AI4Privacy, 25 scenarios for mHealth, and 29 scenarios for PlayStore.

For each taxonomy, we construct a stratified 80/20 split subject to two constraints: (i) both partitions provide full label coverage, such that each label appears at least once in both partitions, and (ii) label distributions are approximately preserved across partitions. The larger partition serves as the \textit{evaluation set} processed by the annotation pipeline, while the smaller partition serves as the \textit{example set} for the RAG mechanism, supplying few-shot demonstrations to the LLM components.\footnote{We use this terminology to avoid confusion with conventional train/test splits, as no statistical model parameters are trained in our pipeline.}

\subsubsection{Models and infrastructure}
\label{subsubsec:models_and_infrastructure}

To reduce coupling between generation and annotation, we employ different LLMs for the two pipelines. Synthetic traffic is generated using \texttt{unsloth/GLM-5-GGUF:Q4\_K\_M}. All annotation stages (classification, annotation, and review) use \texttt{unsloth/Qwen3.5-397B-A17B-GGUF:Q4\_K\_M}, a Mixture-of-Experts model with 397B total parameters and 17B active parameters per forward pass, served via \texttt{llama.cpp}. Inference is executed on a dedicated GPU server equipped with four NVIDIA H200 NVL GPUs.

\subsection{Results}
\label{subsec:results}

As described in Section~\ref{subsec:methodology}, we evaluate the pipeline at five checkpoints corresponding to classification, single-stage annotation, two-stage (targeted) annotation, and the respective reviewed variants. Table~\ref{tab:evaluation_summary} summarises the $F_1$ scores across all checkpoints and datasets (see Appendix~\ref{app:results} for precision and recall). Overall, performance is strongest on \textit{AI4Privacy} and \textit{mHealth}, where label-level detection is near-ceiling already at the classification stage and instance-level extraction remains high across both annotation configurations. In contrast, \textit{PlayStore} is consistently more challenging, indicating that this taxonomy provides a weaker upstream signal and that its value formats and structures are harder to annotate reliably.

At the annotation stage, the two-stage configuration matches or slightly outperforms single-stage annotation on \textit{AI4Privacy} and \textit{mHealth}. The gains are most visible at the instance level, suggesting that conditioning extraction on predicted label sets primarily improves coverage of values rather than changing type detection. On \textit{PlayStore}, the picture differs: two-stage annotation improves label-level detection, but does not translate into a corresponding improvement in instance-level extraction, where single-stage annotation remains marginally stronger. Taken together, these results indicate that the benefit of architectural decomposition depends on the reliability of the classifier and on how readily value boundaries can be identified for a given taxonomy.

Finally, the review stage does not yield consistent improvements. Across datasets, reviewed outputs are comparable to, and in several cases slightly worse than, the corresponding unreviewed annotation outputs. This suggests that under the present operating regime, where annotation quality is already high on two of the three taxonomies, blanket review using the same underlying model provides limited net benefit, and that its utility may depend on selectively targeting messages that are difficult or ambiguous.

\subsection{Discussion}
\label{subsec:discussion}

The results address our research questions while also delimiting the scope of the evidence provided by our synthetic evaluation. Regarding \textbf{RQ1} and \textbf{RQ3}, the pipeline demonstrates that LLM-based annotation can identify and annotate explicitly transmitted PII values in HTTP message bodies when the taxonomy is supplied at runtime. The synthetic, ground-truth annotated HTTP messages further provide a practical substrate for controlled evaluation and exemplar-based prompting. Performance on \textit{AI4Privacy} and \textit{mHealth} indicates that the pipeline is effective when taxonomy definitions align with typical surface forms and contextual cues in HTTP payloads. By contrast, the drop on \textit{PlayStore} shows that robustness is not uniform across taxonomies. We interpret this as evidence that annotation difficulty depends strongly on taxonomy design: broad, policy-facing categories are less directly associated with distinctive value formats than more concrete PII or PHI categories. Since all datasets maintain full label coverage in both the evaluation set and the RAG exemplar pool, these differences are unlikely to stem solely from missing demonstrations; rather, they point to challenges introduced by semantic granularity, contextual dependence, and weak correspondence between labels and observable payload values.

With respect to \textbf{RQ2}, the ablation study shows that architectural decomposition matters, but its benefits depend on upstream classification quality. The two-stage approach yields small but consistent gains where classification is reliable, primarily through higher instance-level recall. However, the \textit{PlayStore} results show that targeted annotation can inherit noisy classifier outputs and trade precision for recall. For broad or technically embedded labels, future variants may benefit from stronger boundary constraints, delimiter-aware post-processing, placement-specific normalisation, or label-specific confidence checks. The review stage, in contrast, provides limited net benefit: across datasets, it corrects some errors but introduces others. One likely reason is that the same model family is used for annotation and review, which may reproduce shared assumptions rather than provide an independent correction signal. This motivates selective or heterogeneous review strategies, such as triggering review only for low-confidence messages, using a different reviewer model, or involving human adjudication for ambiguous cases.

The use of synthetic HTTP traffic enables controlled taxonomy coverage, avoids exposing real user traffic, and provides ground-truth annotations at a scale that would be costly to obtain manually. Manual validation of scenarios, requests, and annotations mitigates obvious structural and semantic errors, but it cannot establish that the generated traffic follows the same distribution as real-world application traffic. Real traffic may contain malformed payloads, undocumented schemas, application-specific encodings, compression, mixed-language values, obfuscation, telemetry artefacts, or transformed identifiers that are underrepresented in the synthetic corpora. The reported scores should therefore be interpreted as evidence of feasibility under controlled conditions, not as final estimates of performance in deployed systems.

The present evaluation is further limited to explicitly transmitted PII values in HTTP request bodies. While the generator and annotation framework can in principle be extended to paths, query parameters, headers, cookies, and response bodies, this remains to be evaluated because these components provide different contextual cues and may require different normalisation and boundary rules. Similarly, the approach targets text-based, recoverable values and does not address encrypted traffic, binary payloads, or PII that is only inferable from combinations of non-PII attributes.

Finally, the absolute scores reflect a single model configuration for generation and annotation. Their magnitude may vary across model families, parameter scales, quantisation levels, prompt variants, and exemplar-selection strategies. LLM-based PII analysis also raises practical deployment concerns, including latency, cost, energy use, and the risk of exposing sensitive values to external providers. We therefore view the pipeline primarily as a flexible annotation and audit-support mechanism, rather than as a direct replacement for lightweight production detectors. We consider the most promising path for future research and deployment to be a hybrid one: using LLM-based annotation to create or refine labelled data under evolving taxonomies, and subsequently distilling these annotations into smaller discriminative models for high-throughput deployment.
\section{Conclusion}
\label{sec:conclusion}

In this work, we presented an LLM-based, taxonomy-agnostic pipeline for annotating explicitly transmitted PII values in HTTP message bodies. The pipeline combines deterministic normalisation, label-level classification, targeted value extraction, and an inference control layer based on RAG and output validation. To mitigate labelled-data scarcity, we further introduced a taxonomy-driven synthetic HTTP message generator that produces ground-truth annotated traffic for controlled evaluation and exemplar-based prompting. Our evaluation shows that the pipeline can identify PII types and extract corresponding values under runtime-specified taxonomies, with strong performance on concrete PII and PHI taxonomies and lower, but still informative, performance on broader policy-facing labels.

These findings highlight that taxonomy-agnostic PII value annotation can support privacy auditing workflows that must adapt to evolving regulatory, organisational, or sector-specific taxonomies without retraining. The ability to re-annotate the same traffic under alternative taxonomies also provides a route towards improved comparability across empirical studies whose label schemes differ in domain focus and granularity. At the same time, our results underline that taxonomy design, payload structure, and boundary fidelity materially affect annotation quality. Future work should therefore validate the approach on real-world traffic captures, including obfuscated or transformed values; extend evaluation beyond request bodies to headers, paths, queries, cookies, and responses; and improve reliability and cost-efficiency through selective or heterogeneous review. We consider the most promising path for future research and deployment to be a hybrid one: using LLM-based annotation to create or refine labelled data under evolving taxonomies, and subsequently distilling these annotations into smaller discriminative models for high-throughput deployment.

\section*{Data Availability and use of Generative AI}

Although generative AI is the main focus of this paper, all content was created manually. We did, however, use generative AI tools to polish the text and fix mistakes. You can access our source code and datasets on GitHub: \url{https://github.com/tomcory/http-pii-annotator}.

\bibliographystyle{plain}
\bibliography{references}

\appendices

\onecolumn

\section{Evaluation Results}
\label{app:results}

\begin{table}[h]
\centering
\caption{Per-step precision (P), recall (R), and $F_{1}$ at label level and instance level (fuzzy and exact) across all three taxonomies.}
\label{tab:results_all_taxonomies_prf1}
\footnotesize
\begin{tabular}{lccc ccc ccc}
\toprule
 &
\multicolumn{3}{c}{Label-Level} &
\multicolumn{3}{c}{Fuzzy Instance} &
\multicolumn{3}{c}{Exact Instance} \\
\cmidrule(lr){2-4}\cmidrule(lr){5-7}\cmidrule(lr){8-10}
Dataset/Stage & P & R & $F_{1}$ & P & R & $F_{1}$ & P & R & $F_{1}$ \\
\midrule

\multicolumn{10}{l}{\textbf{\textit{AI4Privacy}}} \\
\midrule
Classifier         & 0.9779 & 0.9942 & 0.9860 & --     & --     & --     & --     & --     & -- \\
Annotator (single-stage)          & 0.9838 & 0.9877 & 0.9857 & 0.9671 & 0.9769 & 0.9720 & 0.9329 & 0.9425 & 0.9377 \\
Annotator (two-stage) & 0.9836 & 0.9942 & 0.9888 & 0.9685 & 0.9869 & 0.9776 & 0.9348 & 0.9529 & 0.9438 \\
Reviewer (single-stage)           & 0.9797 & 0.9914 & 0.9855 & 0.9583 & 0.9766 & 0.9674 & 0.9231 & 0.9409 & 0.9319 \\
Reviewer (two-stage)  & 0.9812 & 0.9942 & 0.9876 & 0.9650 & 0.9845 & 0.9747 & 0.9306 & 0.9496 & 0.9400 \\
\midrule

\multicolumn{10}{l}{\textbf{\textit{mHealth}}} \\
\midrule
Classifier         & 0.9781 & 0.9936 & 0.9858 & --     & --     & --     & --     & --     & -- \\
Annotator (single-stage)          & 0.9813 & 0.9872 & 0.9843 & 0.9552 & 0.9635 & 0.9593 & 0.9511 & 0.9587 & 0.9549 \\
Annotator (two-stage) & 0.9824 & 0.9938 & 0.9880 & 0.9627 & 0.9781 & 0.9703 & 0.9584 & 0.9733 & 0.9658 \\
Reviewer (single-stage)           & 0.9834 & 0.9907 & 0.9870 & 0.9572 & 0.9627 & 0.9599 & 0.9522 & 0.9574 & 0.9548 \\
Reviewer (two-stage)  & 0.9841 & 0.9922 & 0.9881 & 0.9612 & 0.9682 & 0.9647 & 0.9563 & 0.9632 & 0.9597 \\
\midrule

\multicolumn{10}{l}{\textbf{\textit{PlayStore}}} \\
\midrule
Classifier         & 0.8718 & 0.9538 & 0.9109 & --     & --     & --     & --     & --     & -- \\
Annotator (single-stage)          & 0.9044 & 0.9138 & 0.9091 & 0.8576 & 0.8691 & 0.8633 & 0.8001 & 0.8121 & 0.8061 \\
Annotator (two-stage) & 0.8899 & 0.9488 & 0.9184 & 0.8285 & 0.8995 & 0.8625 & 0.7698 & 0.8354 & 0.8012 \\
Reviewer (single-stage)          & 0.8863 & 0.9251 & 0.9053 & 0.8370 & 0.8780 & 0.8570 & 0.7822 & 0.8222 & 0.8017 \\
Reviewer (two-stage)  & 0.8755 & 0.9425 & 0.9078 & 0.8196 & 0.8945 & 0.8554 & 0.7633 & 0.8346 & 0.7973 \\
\bottomrule
\end{tabular}
\end{table}

\clearpage

\section{Annotator System Prompt}
\label{app:app:prompt_annotator}

\UseRawInputEncoding
\begin{lstlisting}[language=Markdown]
You are a **PII value annotator** for privacy research. Your task is to identify and label **all PII values** that appear in a given **HTTP request body** (payload). The output will be used to evaluate PII classifiers and masking/annotation models.

You will be given:
- A **PII taxonomy table** (authoritative label scheme with `PII Data type` values).
- One HTTP **body** as a raw text string. The body may be JSON, form-encoded (`key=value&...`), multipart metadata (text fields only), XML, logs, plain text, or nonstandard/obfuscated formats.

**Important assumption:** The body is already fully decoded upstream. Do not apply URL decoding or other decoding. Annotate exactly what you see.

---
## 1) Output format
Return **ONLY** a valid JSON object matching this schema:

- Top-level object has exactly one key: `annotations`
- `annotations` is an array of objects
- Each object has exactly:
  - `data_type` (string): the PII data type of the annotated value
  - `value` (string): the exact PII span as it appears in the body

No additional keys are allowed anywhere.  
No commentary, no markdown, no code fences - **output JSON only**.

Example structure (illustrative only):
```json
{
  'annotations': [
    { 'data_type': '<EXACT_TAXONOMY_LABEL>', 'value': '<VERBATIM_SUBSTRING_FROM_BODY>' }
  ]
}
```

---
## 2) Taxonomy adherence
- You MUST use **ONLY** `PII Data type` values that appear in the provided taxonomy table.
- For every annotation, **copy the `data_type` string verbatim** from the taxonomy (exact case, spacing, punctuation). Do not paraphrase.
- Do **not** assume any particular labels exist. The taxonomy may vary between tasks.
- Any examples in this prompt are **illustrative of disambiguation logic only**; they are *not* a guaranteed list of valid labels. If an example label is not present in the provided taxonomy, you MUST NOT use it.
- If a potential PII value does not clearly match any taxonomy `PII Data type`, **do not annotate it**.

---
## 3) What to annotate
Annotate **PII values** present in the body. Values may appear in:
- JSON objects/arrays
- Form-encoded bodies
- XML attributes/text
- Plain text lines
- Nonstandard payloads with unusual separators or obfuscated keys

Annotate the **PII value only**, not the key name or surrounding syntax.

Examples:
- Good: `john.doe@example.com`
- Bad: `email=john.doe@example.com`
- Good: `+49 30 1234567890`
- Bad: `phone:+49 30 1234567890`

Keep punctuation that is inherently part of the value (e.g., `@` in email, `+` in phone, hyphens in UUID-like identifiers).

---
## 4) Span rules
1. **Minimal span**: select the smallest substring that is the PII value (exclude keys, quotes, separators, and surrounding whitespace).
2. **Multi-value lists**: if multiple PII values appear in a list (comma-separated, array, repeated fields), annotate **each value separately**.
3. **Split across fields**: if a combined concept is split into multiple fields, annotate each field's value separately.  
   - Example: `firstname=John&lastname=Doe`: annotate `John` and `Doe` separately (do NOT create a single 'John Doe' value).
4. **Deduplicate repeated values (strong rule)**: deduplicate **globally by exact value string**.  
   - If the **exact same value string** appears multiple times anywhere in the body (even under different keys), include it **only once** in `annotations`.
   - Do not attempt 'semantic deduplication' (e.g., do not merge `John` and `Doe`).
5. **No inference**: annotate only what is explicitly present; do not guess missing pieces or reconstruct values.

---
## 5) Use context to resolve ambiguity
Many values are ambiguous in isolation (especially numbers and short strings). Use **surrounding context**—such as nearby keys, units, structural position, and typical formatting—to choose the correct type **when possible**.

- Only annotate a value if the context makes the type **reasonably unambiguous**.
- If you cannot disambiguate confidently, **do not annotate**.

Examples of *logic* (labels shown are placeholders; use only labels that exist in the provided taxonomy):
- If the key is `sdk_version` or `api_level` and the value is `31`, map it to the taxonomy's API-level-like data type (if present).
- If the key is `age` and the value is `29`, map it to the taxonomy's age-like data type (if present).
- If the key is `weight` and the value includes units like `kg`/`lb`, map it to the taxonomy's body-weight-like type (if present).
- If the key is `tz` and the value looks like `Europe/Berlin`, map it to the taxonomy's timezone-like type (if present).

Counterexample:
- A bare `31` with no key, unit, or context: do not annotate.

---
## 6) Disambiguation guidance
Prefer the **most specific** matching taxonomy type when context supports it. Use value patterns + nearby keys (not guesses). If the taxonomy does not contain a matching specific type, do not 'approximate' with an unrelated label.

If a value could match multiple types and context does not resolve it, **do not annotate**.

---
## 7) Quality checklist
Before returning output:
- Output is valid JSON and conforms exactly to the schema (no extra keys).
- Every `data_type` exactly matches a taxonomy `PII Data type` (verbatim).
- Every `value` is an exact substring from the provided body (verbatim).
- All detectable, context-resolvable taxonomy-matching PII values are included.
- Exact-duplicate values are included only once.

Return only the JSON object.
\end{lstlisting}

\clearpage

\section{Label Taxonomies}
\label{app:label_taxonomies}

\begin{table}[h]
\centering
\caption{AI4Privacy pii-masking-200k label taxonomy~\cite{ai4privacy200k, mainetti2025detecting}.}
\label{tab:pii_examples_sensitivity}
\footnotesize
\begin{tabular}{p{0.34\linewidth} p{0.1\linewidth} p{0.46\linewidth}}
\toprule
PII Data type & Sensitivity level & Description \\
\midrule
Name & medium & Personal name string \\
Email & high & Email address identifier \\
Phone number & high & Telephone number (international/national format) \\
Credit card number & high & Payment card number (PAN) \\
Passport number & high & Passport identifier / MRZ-style string \\
Social security number & high & National SSN-style identifier \\
Date of birth & medium & Birth date (ISO or locale-specific format) \\
Address & medium & Postal address (street, number, city) \\
Zip code & medium & Postal/ZIP code \\
IP address & high & IPv4/IPv6 network address \\
Bank account number & high & Bank account identifier (e.g., IBAN-like) \\
Routing number & medium & Bank routing / sort code identifier \\
Driver's license number & medium & Driver's licence identifier \\
License plate number & medium & Vehicle registration plate \\
Vehicle identification number & low & Vehicle identification number (VIN) \\
Bank account details & high & Structured banking details (account type, sort code, holder) \\
Medical record number & high & Hospital/clinic record identifier (MRN-style) \\
Health insurance number & high & Health insurance member/policy identifier \\
Employee number & medium & Employee identifier \\
Student ID & medium & Student identifier \\
Government ID & high & Government-issued identifier (national ID style) \\
Geolocation & high & Place name or coarse location token (e.g., geohash) \\
Biometric data & high & Biometric template/embedding representation \\
Mother's maiden name & medium & Maiden name token \\
Family member names & medium & Names of relatives/household members \\
Place of birth & medium & Birthplace (city/region) \\
Citizenship & medium & Citizenship code(s) or status \\
Gender & medium & Gender identity value \\
Sexual orientation & medium & Sexual orientation category \\
Race/ethnicity & medium & Race/ethnicity category \\
Nationality & medium & Nationality descriptor \\
Religious beliefs & medium & Religious affiliation / belief category \\
Political affiliation & medium & Political affiliation category \\
Professional license number & medium & Professional licence identifier (e.g., medical/bar) \\
Education history & medium & Education credential/history snippet \\
Employment history & medium & Employment record snippet \\
Income level & low & Income bracket / level indicator \\
Financial status & low & Financial ratio/band indicator (e.g., debt-to-income) \\
Social media handles & low & Social media username/handle \\
Account numbers & high & Customer/account identifier(s) \\
Transaction history & high & Transaction record snippet (amount, merchant, timestamp) \\
Digital signature & high & Cryptographic signature token (e.g., base64) \\
Purchase history & medium & Purchase/order record snippet \\
Subscription information & medium & Subscription plan/status metadata \\
Health information (e.g., conditions, treatments) & high & Health condition/treatment/medication mention \\
Emergency contact information & low & Emergency contact name/relationship/phone \\
Insurance policy number & high & Insurance policy identifier \\
Academic records & medium & Academic performance record (grades/GPA) \\
Tax identification number & low & Tax identifier string \\
License key or serial number & medium & Product licence key / serial number \\
Location data (GPS) & high & GPS coordinates or latitude/longitude pair \\
\bottomrule
\end{tabular}
\end{table}

\clearpage

\begin{table}[h]
\centering
\caption{PII/PHI label taxonomy for mHealth apps (Cory et al. (2024) ~\cite{CoryIWPE2024}).}
\label{tab:pii_taxonomy_examples}
\footnotesize
\begin{tabular}{p{0.15\linewidth} p{0.15\linewidth} p{0.15\linewidth} p{0.45\linewidth}}
\toprule
PII Category & PII Subcategory & PII Data type & Description \\
\midrule
Device or other IDs & Hardware identifiers & Device model & The commercial model name of the user's device hardware. \\
 &  & Screen resolution & The pixel dimensions of the device's display, often reported as width $\times$ height. \\
 & Software identifiers & Carrier name & The name of the mobile network operator or SIM provider. \\
 &  & OS build & The operating system build identifier or version string. \\
 &  & API level & The numeric API level of the operating system (e.g., Android SDK level). \\
\midrule
Location & Approximate location & Locale & The locale setting reflecting language and region preferences. \\
 &  & Timezone & The user's time zone specification. \\
 &  & Country & The country or region name associated with the user's location. \\
 &  & Country code & The ISO country code representing the user's country. \\
 &  & City & The name of the city or town where the user is located. \\
 & Precise location & Latitude & The GPS latitude coordinate, often in decimal degrees. \\
 &  & Longitude & The GPS longitude coordinate, often in decimal degrees. \\
\midrule
Personal info & User identifiers & Name & How a user identifies themselves, such as their first and last name or nickname. \\
 &  & Email address & A user's email address. \\
 &  & Advertising ID & A unique, resettable identifier used for advertising and analytics. \\
 &  & Age & A user's age in years. \\
 &  & Date of birth & The user's birth date, often as day--month--year. \\
 &  & Gender & The gender identity reported by the user. \\
\midrule
Fitness info & Body measurements & Body height & The user's height measurement. \\
 &  & Body weight & The user's current body weight. \\
 &  & Body weight goal & The target weight that the user aims to achieve. \\
 & Fitness info & Step count & The number of steps the user has taken in a defined period. \\
 &  & Eating habits & Information about the user's diet patterns or dietary preferences. \\
 &  & Fitness goals & Specific fitness targets set by the user. \\
 &  & Fitness level & The user's assessed fitness level or self-reported status. \\
 &  & Mental wellbeing & Qualitative information on the user's mental or emotional health status. \\
 &  & Sleep habits & Patterns in the user's sleep, such as average hours or bedtime. \\
\midrule
Health info & Female health info & Menstrual cycle length & The length of the user's menstrual cycle. \\
 &  & Period start date & The date when the user's menstrual period begins. \\
 &  & Period length & The number of days the user's menstrual period lasts. \\
 &  & Period symptoms & Recorded symptoms experienced during menstruation. \\
 &  & Birth control & The contraceptive method used. \\
 & Medical info & Sexual activity & Information regarding sexual behaviour or activity frequency. \\
 &  & Body temperature & Body temperature readings, typically in degrees Celsius or Fahrenheit. \\
 &  & Heart rate & Beats per minute measurement of the user's heart rate. \\
 &  & Blood pressure & Systolic and diastolic pressure readings. \\
 &  & Glucose levels & The user's blood glucose measurement. \\
 &  & Medical conditions & Recorded medical diagnoses or health conditions. \\
\bottomrule
\end{tabular}
\end{table}

\clearpage

\begin{table}[t]
\centering
\caption{Google Play data safety label taxonomy~\cite{datasafety}.}
\label{tab:playstore_taxonomy}
\footnotesize
\begin{tabular}{p{0.18\linewidth} p{0.2\linewidth} p{0.52\linewidth}}
\toprule
PII Category & PII Data type & Description \\
\midrule
Location & Approximate location & User or device physical location to an area greater than or equal to 3 square kilometres, such as the city a user is in, or location provided by Android's \texttt{ACCESS\_COARSE\_LOCATION} permission. \\
 & Precise location & User or device physical location within an area less than 3 square kilometres, such as location provided by Android's \texttt{ACCESS\_FINE\_LOCATION} permission. \\
\midrule
Personal info & Name & How a user refers to themselves, such as their first or last name, or nickname. \\
 & Email address & A user's email address. \\
 & User IDs & Identifiers that relate to an identifiable person. For example, an account ID, account number, or account name. \\
 & Address & A user's address, such as a mailing or home address. \\
 & Phone number & A user's phone number. \\
 & Race and ethnicity & Information about a user's race or ethnicity. \\
 & Political or religious beliefs & Information about a user's political or religious beliefs. \\
 & Sexual orientation & Information about a user's sexual orientation. \\
 & Other info & Any other personal information such as date of birth, gender identity, veteran status, etc. \\
\midrule
Financial info & User payment info & Information about a user's financial accounts, such as credit card number. \\
 & Purchase history & Information about purchases or transactions a user has made. \\
 & Credit score & Information about a user's credit score. \\
 & Other financial info & Any other financial information, such as user salary or debts. \\
\midrule
Health and fitness & Health info & Information about a user's health, such as medical records or symptoms. \\
 & Fitness info & Information about a user's fitness, such as exercise or other physical activity. \\
\midrule
Messages & Emails & A user's emails, including the email subject line, sender, recipients, and the content of the email. \\
 & SMS or MMS & A user's text messages, including the sender, recipients, and the content of the message. \\
 & Other in-app messages & Any other types of messages. For example, instant messages or chat content. \\
\midrule
Photos and videos & Photos & A user's photos. \\
 & Videos & A user's videos. \\
\midrule
Audio files & Voice or sound recordings & A user's voice, such as a voicemail or a sound recording. \\
 & Music files & A user's music files. \\
 & Other audio files & Any other user-created or user-provided audio files. \\
\midrule
Files and docs & Files and docs & A user's files or documents, or information about their files or documents, such as file names. \\
\midrule
Calendar & Calendar events & Information from a user's calendar, such as events, event notes, and attendees. \\
\midrule
Contacts & Contacts & Information about the user's contacts, such as contact names, message history, and social graph information like usernames, contact recency, contact frequency, interaction duration and call history. \\
\midrule
App activity & App interactions & Information about how a user interacts with the app. For example, the number of times they visit a page or sections they tap on. \\
 & In-app search history & Information about what a user has searched for in your app. \\
 & Installed apps & Information about the apps installed on a user's device. \\
 & Other user-generated content & Any other user-generated content not listed here, or in any other section. For example, user bios, notes, or open-ended responses. \\
 & Other actions & Any other user activity or actions in-app not listed here, such as gameplay, likes, and dialogue options. \\
\midrule
Web browsing & Web browsing history & Information about the websites a user has visited. \\
\midrule
App info and performance & Crash logs & Crash log data from your app. For example, the number of times your app has crashed, stack traces, or other information directly related to a crash. \\
 & Diagnostics & Information about the performance of your app, such as battery life, loading time, latency, framerate, or any technical diagnostics. \\
 & Other app performance data & Any other app performance data not listed here. \\
\midrule
Device or other IDs & Device or other IDs & Identifiers that relate to an individual device, browser or app. For example, an IMEI number, MAC address, Widevine Device ID, Firebase installation ID, or advertising identifier. \\
\bottomrule
\end{tabular}
\end{table}

\clearpage

\end{document}